# Cloud-Based Hierarchical Imitation Learning for Scalable Transfer of Construction Skills from Human Workers to Assisting Robots

Hongrui Yu[1], Vineet R. Kamat[2*], Carol C. Menassa[3]

[1] Ph.D. Candidate, Department of Civil and Environmental Engineering, University of Michigan. E-mail: yhongrui@umich.edu

[2*] Professor, Department of Civil and Environmental Engineering, University of Michigan. E-mail: vkamat@umich.edu

[3] Professor, Department of Civil and Environmental Engineering, University of Michigan. E-mail: menassa@umich.edu

**Abstract**

Assigning repetitive and physically-demanding construction tasks to robots can alleviate human workers' exposure to occupational injuries that often result in significant downtime or premature retirement. Yet, the successful delegation of construction tasks and achieving high-quality robot-constructed work requires transferring necessary dexterous and adaptive artisanal construction craft skills from workers to robots. Predefined motion planning scripts tend to generate rigid and collision-prone robotic behaviors in unstructured construction site environments. In contrast, Imitation Learning (IL) offers a more robust and flexible skill transfer scheme. However, the majority of IL algorithms rely on human workers to repeatedly demonstrate task performance at full scale, which can be counterproductive and infeasible in the case of construction work. To address this concern, this paper proposes an immersive, cloud robotics-based virtual demonstration framework that serves two primary purposes. First, it digitalizes the demonstration process, eliminating the need for repetitive physical manipulation of heavy construction objects. Second, it employs a federated collection of reusable demonstrations that are transferable for similar tasks in the future and can thus reduce the requirement for repetitive illustration of tasks by human agents. Additionally, to enhance the trustworthiness, explainability, and ethical soundness of the robot training, this framework utilizes a Hierarchical Imitation Learning (HIL) model to decompose human manipulation skills into sequential and reactive sub-skills. These two layers of skills are represented by deep generative models, enabling adaptive control of robot actions. The proposed framework has the potential to mitigate technical adoption barriers and facilitate the practical deployment of full-scale construction robots to perform a variety of tasks with human supervision. By delegating the physical strains of construction work to human-trained robots, this framework promotes the inclusion of workers with diverse physical capabilities and educational backgrounds within the construction industry.

**Introduction**

The construction industry currently faces substantial challenges in its workforce (Delvinne et al. 2020). An estimated labor shortage of 430,000 construction workers (BLS 2023) is compounding the industry's difficulties, which are further amplified by long-forecasted projections of increasing demand (Groves et al. 2013; AGC 2016; Delvinne et al. 2020). Notably, for labor-intensive construction work, this shortage directly results in project delays and increased financial costs (Sokas et al. 2019). According to RSMeans, the burgeoning labor shortages have contributed to a 10% increase in total construction costs in 2016 (RSMeans 2016).

Construction robots can play a crucial role in addressing the workforce gap and mitigating the challenges posed by labor shortages (Lundeen et al. 2018; Wang et al. 2021; Brosque et al. 2022). Robots possess superior physical capabilities and thus excel in handling heavy and repetitive construction tasks (Liang et al. 2021; Chen et al. 2022). Furthermore, they are less prone to physical fatigue and cognitive lapses (Shayesteh & Jebelli 2021; Lee et al. 2022). By deploying construction robots, significant improvements can be achieved in construction productivity, leading to reduced delays and construction costs (Ryu et al. 2020; Pan and Pan 2020).

*Robot Craft Skill Learning*

However, unstructured construction site environments pose inherent challenges for robots when performing construction tasks (Chen et al. 2016; Liang and Cheng 2023). For construction robots, accomplishing any complex motion planning process requires intricate motion control scripts (Cai et al. 2022; Zhu et al. 2022). Modeling and learning adaptive construction skills become increasingly challenging due to the abundance of information influencing the reasoning behind specific actions (Makondo et al. 2015; Xu et al. 2020). For example, in Lundeen et al. (2017), the geometric configuration of the workpiece was reconstructed by careful calibration of sensors to establish the correlation of choice of robot actions to adapt to the environmental observations. Ignoring these motivations and accompanying spatial context tend to limit the robot to generating inflexible and inadequate motions (Maddadin et al. 2017). In a dynamic and unstructured environment like a construction site, relying solely on rigid robot control schemes significantly increases the risk of getting stuck (i.e., stalling) or colliding with adjacent objects or human workers (i.e., interfering) (Zhu et al. 2022; Sun et al. 2023).

In contrast, human apprentices (junior construction workers) acquire dexterous manipulation skills and complex construction techniques organically by observing experienced workers (Sings et al. 2017; Liang et al. 2021). Through these observations, apprentices instinctively establish correlations between sequential actions and the underlying motivations that drive them. These motivations are influenced by contextual information derived from the environment, previous actions, and overall work progress (Yu et al. 2018; Liang et al. 2021; Lee et al. 2022; Huang et al. 2023). Consequently, workers naturally transform their observations into a cohesive set of actions guided by environmental cues and the sequence of preceding actions.

*Skill Transfer from Human to Robots with Imitation Learning*

Replicating the learning process of humans, robot Imitation Learning (IL) (i.e., Learning from Demonstration) models offer a flexible and effective approach to motion control. These models employ neural networks or sequential chains to map complex environmental observations to desired actions (Liang et al. 2021; Huang et al. 2023). For instance, a deep neural network can represent expected robot action policies based on various factors such as observed equipment status, worker states, task progress, and workpiece state. Reinforcement Learning (RL) is also often employed to simulate the sequential decision-making processes of human workers, forming the prevalent Deep Reinforcement Learning (DRL) framework (DelPreto et al. 2020).

*Low-Workload and Ethical Imitation Learning*

However, applying IL to construction robots presents additional challenges for two primary reasons.

First, construction materials tend to be heavy, and repetitive manipulation during demonstrations can impose increased physical strain on construction workers. To address this issue, this paper proposes a high-fidelity Digital Training Environment for Construction (DTEC) that enables human workers to naturally demonstrate installation tasks by performing construction tasks with Digital Twin models in Virtual Reality (VR). The DTEC connects with Building Information Modeling (BIM) modules, which are capable of being automatically synchronized with site conditions (Fang and Cho 2015; Du et al. 2018; Wang et al. 2023). This interface also helps increase the worker's trust in construction robots and self-efficacy (Adami et al. 2022).

Additionally, this study proposes a federated construction skill cloud database that can leverage previous knowledge and demonstrations collected from diverse task environments, to reduce the number of required future demonstrations. This federated data collection allows for crowdsourced demonstrations from workspaces representing various temporal and spatial contexts, enabling continuous learning and enhancing scalability in skill transfer between human workers and assisting robots (DelPreto et al. 2020).

Second, considering the ethical and responsible aspects of Artificial Intelligence (AI), the traditional RL-based structure lacks transparency and explainability (Gunning et al. 2017). From a human perspective, the robot learning model appears as a black box, lacking a logical description of learning dynamics and causal relationships. This lack of transparency can increase the cognitive load on construction workers without programming expertise and reduce their trust and willingness to collaborate with robots (Gunning et al. 2017; Park et al. 2023). To address this issue, an explicable construction task decomposition framework is proposed in this paper. It involves decomposing construction skills into sequential and reactive skills. A Hierarchical Imitation Learning (HIL) model is employed to ground the decomposed skills and translate them into robot control instructions. With improved explainability, the robot learning model will appear more trustworthy and easier to use for human workers without programming expertise (Kilian 2020). In the long run, it will result in a more sustainable human-robot relationship and enhance the inclusiveness of workers with diverse educational backgrounds.

Figure 1 summarizes this paper's technical contributions in enabling skill transfer from human workers to assisting construction robots, which include:

1) An immersive virtual demonstration environment, i.e., DTEC, comprising a Digital Twin, Virtual Reality (VR), and wireless data communications integrated with the Robot Operating System (ROS). This training environment enables automated capture of workers' motions and records them as demonstrations for robots, minimizing the physical workload on workers.

2) A cloud robotics framework designed to store, process, and reuse the demonstrations collected from the immersive virtual training environment across diverse task scenarios. This federated framework reduces the necessity for new demonstrations, thereby enhancing the scalability and sustainability of construction robot learning.

3) An explainable representation and transfer scheme for construction skills that effectively translates human manipulation motions into a hierarchical robot learning model capable of decoding both reactive and sequential skills. By incorporating this knowledge back into the cloud database for storage, the scalability and reusability of construction robot programming can be further amplified.

The rest of this paper is organized as follows. First, we provide an introductory literature review of the novel concepts of HIL and cloud robotics, leading to its suitability for application in the construction industry. Second, we present the methodology for the demonstration collection and skill decomposition system based on cloud robotics. Third, we conduct experiments to evaluate the performance of the system, focusing on its ability to reduce the human effort required in both the demonstration and programming phases.

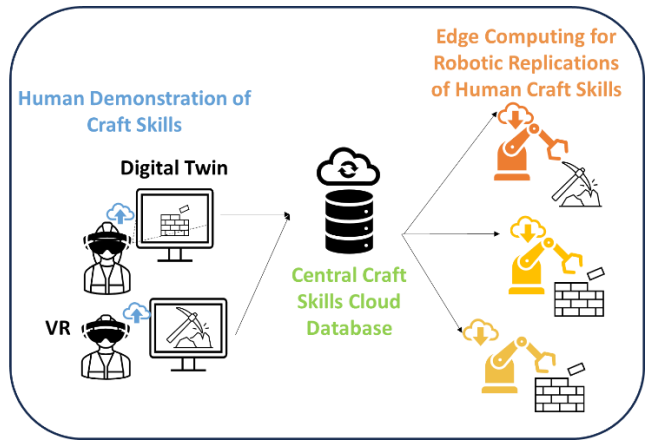

**Figure 1 Cloud-Based Imitation Learning Framework for Construction Robots**

Literature Review

The transfer of skills between human workers and robots has been widely studied to enable robots to adapt to diverse construction environments (Lundeen et al. 2018; Yu et al. 2023; Duan et al. 2023). While pre-defined motion scripts have been considered viable solutions, such as for bricklaying tasks (Wos and Dindorf 2023), such an approach is unable to accommodate the significant variations in the environment and the corresponding adjustments needed in robot actions (Shi et al. 2023). To overcome these limitations, there has been a growing focus on leveraging deep learning and probabilistic models to mimic the adaptive decision-making process of humans in response to environmental observations.

As these models evolve to become more accessible to human workers, there is an increasing demand for scalability, ease-of-use, and explainability (Gunning et al. 2017; Park et al. 2023). The framework proposed in this paper addresses these requirements through the integration of cloud robotics and a hierarchical human skill representation scheme. This section provides an overview of both methods, highlighting their potential application in the construction field. At the end of each sub-section, a summary of the current research status quo and identification of knowledge gaps pertaining to the effective implementation of these concepts on construction sites is presented. This information establishes the motivation for the proposed framework, which is subsequently described in the Methodology section.

*Explainable Human-Robot Master-Apprenticeship*

The seamless transfer of skills between construction workers and robots is essential for achieving high-quality co-robotized construction tasks (Xu et al. 2020; Shi et al. 2023). Construction workers possess dexterity and specialized expertise, enabling them to avoid collisions and effectively address construction-related challenges (Liang et al. 2021). However, given that programming and robotics training are not typically included in their current skill set, programming robots becomes an exceedingly difficult task (Liu et al. 2019). Liang et al. (2020) proposed the construction robot IL (i.e., Learning from Demonstrations) as an approach to address this problem. This approach to Human-Robot Collaboration requires human workers to demonstrate the natural execution process of tasks. These demonstrations are captured by video recording and then are correlated with environmental observations using a DRL model. However, human workers must repetitively manipulate heavy construction materials to provide sufficient demonstration data. To address this problem, Huang et al. (2022) explored the use of digital demonstrations to alleviate the physical strain associated with manipulating heavy construction materials. Even though this study creatively improved the demonstration environment, its kinesthetic demonstration, involving the human moving the robot end-effector in a VR environment to showcase the desired trajectory, heavily depends on the robot's specific structure, limiting its generalizability to different robot types. To extend the application range of robots, this study designed demonstrations to represent the installation trajectories of tasks rather than the joint states of a specific robot.

Furthermore, as mentioned in the *Introduction* section, a common issue with robot learning models is the lack of transparency and explainability. This lack of clarity and understanding can lead to reduced trust and perceived validity among human workers (Gunning et al. 2017). HIL addresses this concern by mimicking the natural way in which humans break down and simplify problems (Zhang et al. 2021). In HIL, the high-level planning serves the purpose of task decomposition and connects the sequence of the decomposed tasks (Abdo et al. 2012). The low-level policies link task execution steps with environmental observations such as task progress, object states, and robot states (Fan et al. 2020). For example, Zhang et al. (2021) decoded the robot pouring tasks into three hierarchies of phase, state, and action to improve task performance, adaptability, and manipulability. Wang et al. (2021) used HIL to teach robots to perform low-clearance insertion assembly tasks with increased sample efficiencies.

By employing two levels of policies, HIL also achieves enhanced learning efficiency and reduces the need for a large number of demonstrations in long-horizon and complex tasks due to compounding errors (Kase et al. 2020; Yu et al. 2018; Ross et al. 2011). For example, Hayashi et al. (2022) reduced 20% of training time by decomposing tasks to a planning layer and a motion primitive layer to ground the elemental motions. There are additional advantages. First, HIL reduces errors and uncertainties accumulated during the robot behavior cloning process, thus improving learning performance (Zhang et al. 2021). Second, it enables the use of conditional probabilities from previously observed actions to enhance confidence and simplify inference of the current state from observations, thereby mitigating perceptual aliasing—where similar environmental observations can lead to confusion between different task steps that serve distinct functions (Kase et al. 2020). For instance, when a robot aims to install three drywall panels consecutively and the environmental observations indicate that the robot is at the location of the second board, there are two possible action options depending on the task progress. By knowing the status of the first board installation, the decision-making process becomes considerably more straightforward.

In the context of construction problems, hierarchical task decomposition has been proposed and explored in some prior work. For instance, Wu et al. (2022) proposed a four-level hierarchy to decompose bricklaying tasks into two levels of subtasks: activities and actions. However, there is limited research available that connects such knowledge with robot learning models. It is still unclear how to decompose construction tasks and clearly represent them with deep robot learning models for robot control. To address this problem, this paper explores how to decompose tasks with two layers: a layer of sequential actions linking the different subtasks and a layer of reactive skills that correlates the environmental observations with choices of subtasks. The details of this technical approach are described in the Methodology section.

*Skill Transfer Scalability with Cloud Robotics*

With HIL, humans can seamlessly transfer skills to robots. However, the current state of robot Imitation Learning (IL) schemes remains primarily need-based, requiring human operators to program robots from scratch whenever a new task arises. Such repetitive programming activities impose an additional workload on human workers. To address this challenge, a common approach is to employ a knowledge database for the sustainable reuse of programmed instructions. For instance, Karp et al. (1994) proposed building a robot knowledge storage database and utilizing a Database Management System (DBMS) to improve communication efficiency. Bandera et al. (2010) introduced a robot knowledge database architecture that combines perceptive information, reflexive behaviors, and known actions.

Compared to local databases, cloud-based databases offer higher convenience for sharing and communication between multiple agents. As cloud-based systems gain popularity, more studies are being conducted to explore effective database architectural designs to adapt to different tasks. Perceptual information, human demonstrations, and robot executions are commonly stored. For dexterous tasks, Hsiao and Lozano-Perez (2006) recorded keyframes of successful grasp trajectories to teach robots precise grasp tasks with an accuracy of 92%. Zhang et al. (2021) illustrated how robots can be instructed to perform pouring tasks using demonstrations that include background information, task objects, and task planning policies.

Besides human-demonstrated trajectories, supervisory commands can also serve as demonstrations and be stored in the robot dexterous manipulation learning database. Yamada et al. (2001) recorded natural language instructions from humans and the corresponding robot motion trajectories, manipulation objects, and elementary motions for a domestic robot operating under human supervision. Rokossa et al. (2013) employed a knowledge database to store analyzed parameters of movement commands, entire motion sequences, velocities, accelerations, blending behaviors, and signal-switching points.

Furthermore, for the hierarchical decomposition of more complex tasks, Kyrarini et al. (2019) proposed storing atomized tasks, associated object states, and learned Gaussian mixture models (GMMs) representing trajectories. Liu et al. (2022) collected both human-demonstrated trajectories and robot execution joint states, gripper dimension parameters, and gripper pose data to teach robots assembly and insertion tasks, effectively reducing the workload for humans. Wang et al. (2020) used stored dual-arm robot pose and point cloud models to enhance the perception and improve the model rebuilding accuracy for assembly tasks.

However, despite these initial studies, it remains unclear what data or demonstrations are needed to compose a feasible cloud database capable of supporting local robot task execution decision-making and planning for both heavy and dexterous construction tasks. Moreover, there is limited knowledge in understanding the impact of the data stored in the cloud database on future task executions in unexpected environments. In this paper, we propose the dual storage of knowledge and raw demonstrations, which is tested and demonstrated using a construction task case study. The previously stored data is also added to the robot learning model for new task situations to investigate its usefulness. The proposed approach is further described in the *Methodology* section.

*Cloud Robotics in Construction*

In the civil sector, cloud databases and computing are widely used for natural disaster responses, worker safety management, waste minimization, building management, and project management informatics (Jiao et al. 2013; Rawai et al. 2013; Balaji et al. 2016; Xu et al. 2018; Wan et al. 2020; Bello et al. 2021; Kohler et al. 2022; Deng et al. 2023). Nonetheless, its application in robotic construction has been relatively limited. Considering the effectiveness cloud databases possess in exchanging information, building a cloud robotics scheme for construction robot training offers significant promise. This paper proposes a federated construction robot learning scheme to leverage previously retrieved demonstrations and information. With edge computing and HIL, the local robot control scripts can also be retrieved on an as-needed basis to optimize computation resources efficiency. The proposed approach can reduce both workers' physical demonstration workload and mental workload when collaborating with robots.

**Research Methodology**

This paper introduces a novel federated learning framework for construction robots, utilizing cloud robotics technology. The framework employs an immersive VR demonstrations interface, which is connected to a cloud database housing both crowdsourced raw demonstration data and explanatory knowledge on hierarchical construction skill decompositions. Within the cloud edge, a Hierarchical Imitation Learning algorithm is utilized to decode and model human adaptive skills from the demonstrations and data, which are subsequently replicated to guide the operations of construction robots. The subsequent sections elaborate on the technical details of each component. In addition, a case study of ceiling installation was used to illustrate how the learning algorithms can be applied to construction tasks.

*Cloud-Robotics Learning Workflow Overview*

Introducing new methods for robotic construction and robot programming inevitably brings changes to the construction practice and workflow. The proposed workflow adapting to the cloud-robotics scheme is illustrated in Figure 2.

In this scheme, the robot starts by recognizing task and target information by scanning AprilTag markers and Building Information Modeling (BIM) database to retrieve the task object's dimensions, material specifications, and locations (Aryan et al. 2021; Wang et al. 2023). Subsequently, the robot accesses related demonstrations available in the cloud database. If sufficient data exists, the robot downloads it to the edge node and commences imitation learning to derive expected installation trajectories.

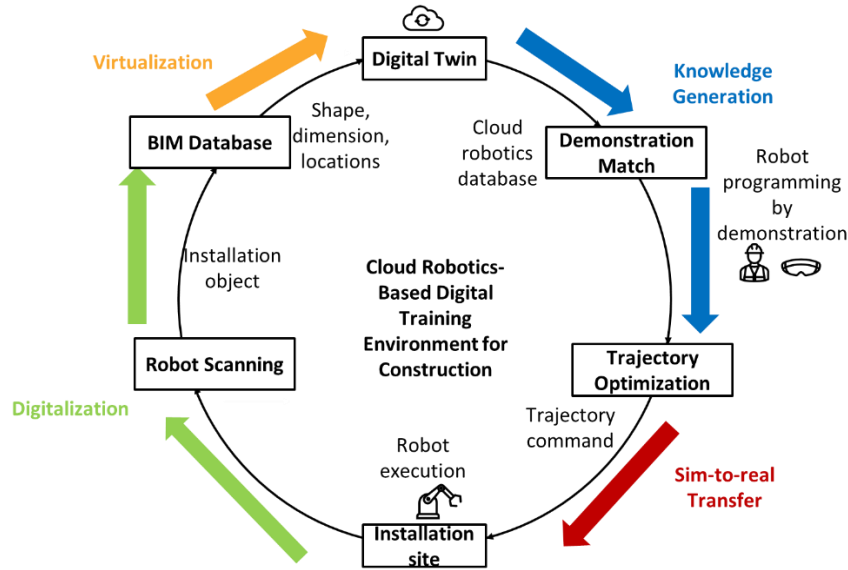

**Figure 2 Proposed Workflow for Cloud-Robotics-Based Construction Learning Scheme**

If cloud data is insufficient, the target information is visualized simultaneously the DTEC to create a digital training setting for human workers to generate demonstrations. Relevant demonstrations from the cloud database, with heterogeneous tasks with similar object locations or geometric configurations, are also downloaded and used in the imitation learning database on the edge node. This process equips the robot with the capability to utilize the gathered information effectively. A hierarchical construction skill modeling system composed of generative and probability model will then be used to replicate the craft skills. Subsequently, based on the robot's type, the edge node then employs Inverse Kinematics (IK) (Tolani et al. 2020), robots of various kinds can then calculate the intended robot joint states and execute the learned trajectory efficiently (Nakanishi et al. 2020). The edge node then utilizes the wireless communication method for real-time robot control based on the IK-calculated joint states.

*Immersive DTE and Demonstration Recordings*

The proposed technical contribution starts with a digital robot training environment (the DTEC) where the workers can teach by naturally performing construction tasks. As mentioned in the *Introduction* section, the repetitive manipulation of heavy construction materials, even for demonstration-only purposes, will cause high physical strain for construction. This framework proposes to perform construction tasks in the virtual environment and remove the repetitive physical interactions with the heavy construction material. This virtual environment has three components: 1) natural task demonstration capture: the motion capture starts with a hand motion tracking system with *HTC VIVE Tracker 3.0* (tracker) motion sensor, which captures the human's hand motions continuously to seize the demonstrated trajectories; 2) trajectory recording: the recorded trajectory will be sent to Robot Operating System (ROS) where a Linux shells script will automatically launch the Rosbag file recorder and give sound notifications on the different stages of demonstration collection, which is also easily connected with the cloud database; 3) digital twin: which can be readily synchronized with the real-time on-site conditions with computer vision approaches, such as AprilTag-based localization and scan-to-BIM digital reconstruction (Kim and Cho 2017; Feng et al. 2021; Wang et al. 2023) and 4) VR: a high-fidelity environment to accommodate the digital twin and human demonstrated motions. The technical approach details of each component are shown as follows.

For an immersive and site-synchronized VR environment, this work leverages the start-of-the-art system (Wang et al. 2023) display the target object's and installation objects' location, texture, and status in VR with low latency and high fidelity. On its basis, box colliders are added to the digital twin objects with dimensions 1 cm larger than the original model. The colliders will stop any potential collisions from 1 cm away. The low-quality demonstrations with collisions can thus be easily identified and avoided.

In addition, a hand motion tracking system was added. The hand tracking system is based on the base station, dongle, a tracker as shown in Figure 3. The base station will emit infrared lasers and the sensors on the tracker will receive this signal for localization. As the localization mechanisms remain non-transparent as a black box (Bauer et al. 2021), experimental testing was used to validate the localization accuracies, as shown in the Experimental Case Study section.

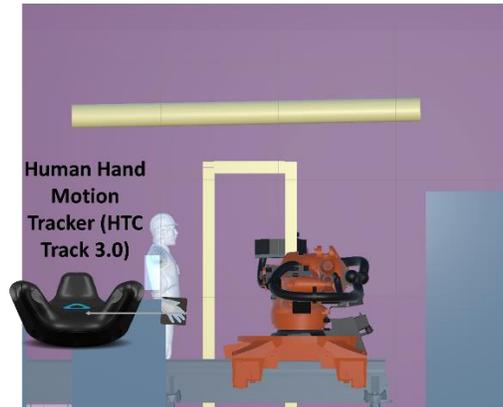

**Figure 3 Human Hand Motion Monitoring with Tracker 3.0**

To reflect the human hand motion, the motion sensor was attached to the human hand in the palm area. When the workers move their hands for demonstrations in the real world, the tracker's pose in VR will be the same as the construction materials in real installation processes. The manipulation objects in VR were set as child objects of the tracker. With such a setting, any motion captured by the tracker will be seamlessly synchronized to the object. The hand motion for manipulating objects can thus be replicated in VR accurately.

Moreover, to ensure applicability across different robot types, this paper proposed to use object trajectories as demonstrations. In the demonstration process, the VR system captures ideal and expected object trajectories required to accomplish construction tasks. By employing IK, the desired joint states necessary to execute the ideal trajectory can be calculated.

*Cloud Robotics for Demonstration Storage and Information Flow*

A cloud database was used to store human demonstrations and ensure convenient access without spatial and temporal limitations. Considering the ease of use, Amazon Web Service (AWS) S3 package was employed. The information flow of the whole system is shown in Figure 4. First, the demonstrations will be transmitted to ROS with ROS# protocol. Second, a Linux shell script automatically saves these trajectories as Rosbag files and use the AWS S3 functions will upload these demonstration files to the cloud. These unorganized cloud data will form a Cloud Data Lake in the AWS cloud database. When the robot needs certain knowledge or information from the cloud, the robot will download the data from the cloud database using HTTP protocols.

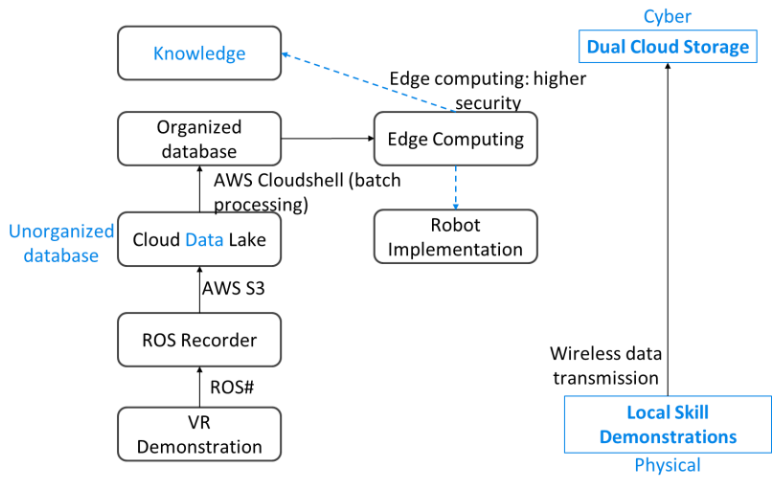

**Figure 4 Information Flow in the Cloud-Robotics Framework**

A dual storage scheme was introduced to sustainably utilize the computational output. As shown in Figure 4, first, the extracted knowledge is utilized to guide on-site robot installations, thereby enhancing the success rate of task execution. Second, the analyzed human motion patterns and knowledge are sent back to the cloud database in both Python pickle models and .csv files and stored for potential applications in future robot installation tasks with similar geometrical configurations. By combining the usage of extracted knowledge with the storage of motion patterns in a cloud database, the system becomes more robust and capable of handling future installation challenges effectively.

*Federated Robot Learning and Data Collection*

With the proposed dual storage scheme, the knowledge and demonstration data can be saved in the cloud database and be used at any time for any task. To alleviate the burden of repetitive demonstrations, this paper also proposes to generalize previous learned knowledge with the collection of demonstration data from heterogeneous tasks, such as tasks from various locations and installation targets of diverse geometric configurations. This heterogeneous and crowdsourced demonstration data naturally forms a federated data collection scheme.

For the newly encountered task, the trajectory will be transformed to the new task scenario by twisting the pose (X,Y,Z) with the task parameter variations, as shown in Eq. 1:

$$p' = \begin{pmatrix} x' \\ y' \\ z' \\ 1 \end{pmatrix} = T \cdot p = \begin{pmatrix} a_{11} & a_{12} & a_{13} & a_{14} \\ a_{21} & a_{23} & a_{23} & a_{24} \\ a_{31} & a_{32} & a_{33} & a_{34} \\ 0 & 0 & 0 & 1 \end{pmatrix} \cdot \begin{pmatrix} x \\ y \\ z \\ 1 \end{pmatrix} \quad (1)$$

$with\ T: task\ parameter\ transformation\ matrix;$

$$a_{11} := x^{new} - x^{old}; a_{22} := y^{new} - y^{old}; a_{11} := z^{new} - z^{old}$$

The trajectory demonstrations from diverse task locations can thus be transformed to have the same target location and concatenated together with the new ad-hoc demonstrations to form an expanded database. The resultant robot learning model performance with expanded dataset is illustrated in the *Experimental Results* section.

*Construction Task Decomposition and Hierarchical Modeling*

With demonstration datasets prepared, the next step is the hierarchical decomposition and learning of construction skills. In traditional construction master-apprentice relationships between human workers, the acquisition of construction skills heavily relies on muscle memory, practical experience, and hands-on work exposure (Sing et al. 2017). In the domain of robot learning and programming, one responsible and explainable approach to replicate such apprenticeship is to motion primitives to decompose a complex, long-horizon task into multiple elemental motions (Schaal et al. 2003), as shown in Figure 5. For example, the ceiling installation task can be broken down to the 8 elemental steps in Figure 6.

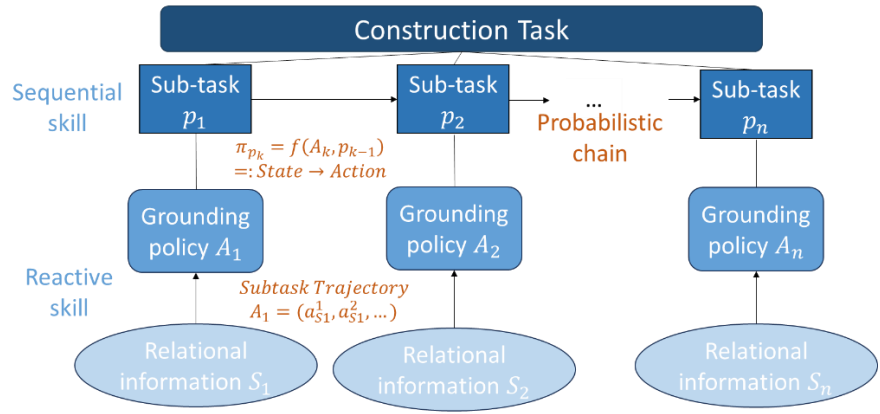

**Figure 5 Construction Task Decomposition**

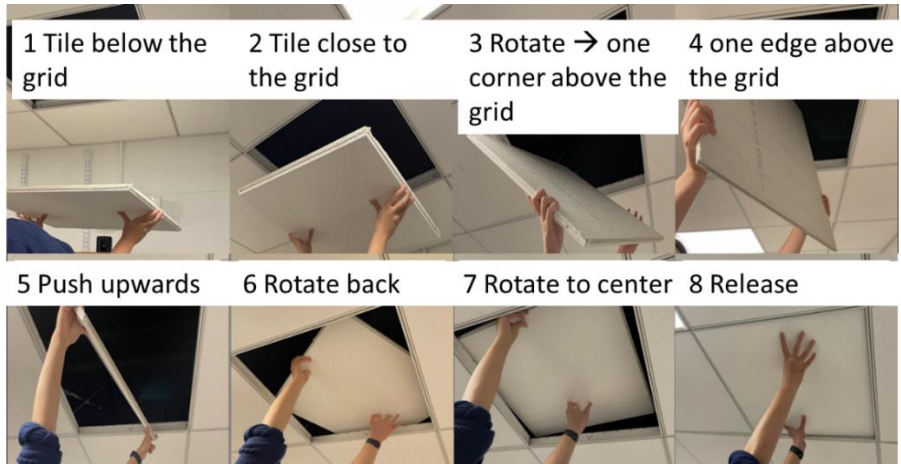

**Figure 6 Motion Primitives for Ceiling Installation Tasks**

As shown in Figure 6, each motion primitive represents an atomic motion or a sub-task that forms the fundamental unit of the overall task. The entire task is treated as a probabilistic chain that connects each step of motion primitives. Each motion primitive also reflects the human worker's reaction to the relational information observed from the environment, such as the distance of picked-up material to the target location. To categorize this decision-making process more comprehensively, the authors propose hierarchical construction skill models comprising *sequential skills and reactive skills*. The details of these two skills are illustrated as follows.

*Sequential Skills*

The sequential skill is proposed to avoid perceptual aliasing, which conditions the choice of actions to the previous history of actions and avoid confusion with similar state observed for different subtasks. With previous actions forming the conditional probability, the probability of mislabeling the environmental observations to those in other subtasks will be largely reduced.

However, the preliminary condition of the probability chain is that the robot agent remembers the previous choice of actions, which implies the necessity of memory capability in the corresponding computation model. A Gated Recurrent Unit (GRU) suits this purpose with the capability to store past action history and use it to update the next-step action decisions. As seen in Figure 7, its gate will use both the previous hidden state and the current input (the actions taken in the previous state) to decide the output of actions. Therefore, it connects past actions with current and future action decisions.

In addition, the authors tested some other models that also condition the current action based on previous actions, such as the Hidden Markov Model (HMM) for high-level sequential skill modeling. As GRU significantly outperforms HMM in terms of learning performances, this paper will only illustrate task decomposition with GRU for sequential skills.

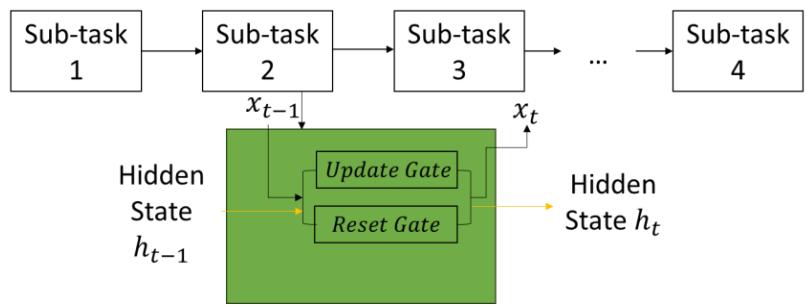

**Figure 7 High-Level Construction Task Decomposition with Sequential Skills**

At each Update and Reset Gate, the input (the observed last action) and hidden state will be passed the activation function to co-determine the next step of action, as shown follows:

$$R_t = \sigma(X_{t-1}W_{t1} + H_{t-1}W_{t2} + b_r)$$
$$U_t = \sigma(X_{t-1}W_{t3} + H_{t-1}W_{t4} + b_u)$$

Where,

$R_t$ and $U_t$ are the Reset and Update Gates at time t, respectively;

$X_{t-1}$ and $H_{t-1}$ are input data of previous action histories and hidden variables;

$W_{t1}, W_{t2}, W_{t3}, W_{t4}, b_r$, and $b_u$ are weight parameters;

*Reactive skills*

The reactive skills represent how human actions are influenced by environmental observations. The observed object state $(A)$ is mapped to a motion primitive $(S)$ through a generative variational autoencoder (VAE) (Kingma and Weeling 2013) model, $A = f(S)$, where the VAE function $f$ is shown in Figure 8.

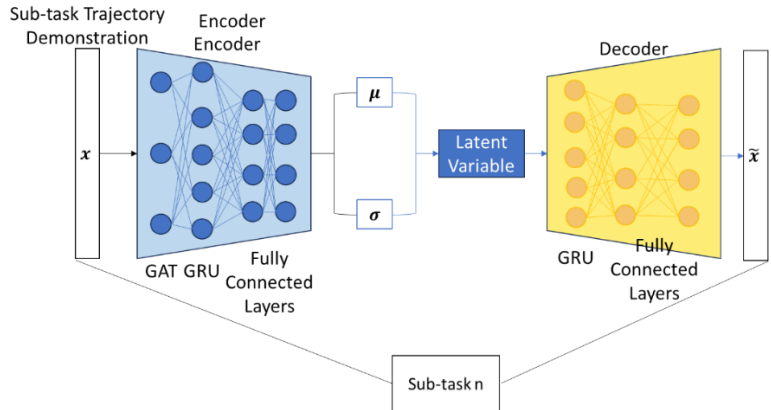

**Figure 8 Low-Level Reactive Skill to Ground Subtask Action Policies**

The input of the lower level of subtask learning is composed of two parts: The first part is the task parameters. For the case study of ceiling installations, we chose the model dimension and location of the ceiling grid to connect with the task information stored in the cloud BIM database. This information will be used to pre-process and prepare the demonstration datasets. The second part of the data input for the VAE model is the installation object state. As this choice has achieved satisfactory results and the goal of this paper is to provide a working system, we have not compared the two options in this study.

**Evaluation Case Study**

*Ceiling Installation Skill Modeling and Transfer*

To evaluate the proposed approach, a case study was conducted to transfer ceiling installation skills to the robot. As described in Liang et al. (2020), ceiling installation is one of the construction tasks that have the highest requirements on dexterity and environmental adaptations. Ceiling tiles are generally larger than the slots in the ceiling grid and need some manipulation to be suspended on a grid. The majority of motion planning algorithms focus on avoiding the obstacles, instead of maneuvering through them, and thus cannot be used for ceiling installation tasks (Liang et al. 2020; Yang et al. 2023).

By observing one experienced construction worker practice the ceiling installation task, eight elemental motions have been summarized, as shown in Figure 6. A graduate student observed the installation and replicated it in VR for the repetitive demonstration data collection. For the next step of computations, a total of eight VAE models will be trained for each elemental motion to represent the reactive skills. A GRU model will be used to connect them and show the sequential skills (Xie et al. 2020).

The object is a standard two by two ceiling grid which is the most commonly used ceiling tile size. The upper left point of the grid has a coordinate of (2.95,1.5,2). Five demonstrations are collected in DTEC to train the robot to perform this task (Dataset 1). Before the task, there were ten more demonstrations performed by the same person stored in the cloud database that are from 1) an additional five from the same location for a 1 by 1 ceiling tile (Dataset 2, a total of 10 demonstrations); and 2) five more from one nearby location of (3.5,2,2) (Dataset 3, a total of 15 demonstrations). The computational results are shown in the *Results* section.

A KUKA KR120 robot was used to execute the task, which is a 6DOF robot with a workload of 120 kg. In this preliminary simulation robot experiments, the robot exhibits some jitter. The jittery behavior was caused by excessively small increments in the demonstrated and learned trajectories. To solve this issue, an adaptive window filter with the window size of 0.1 was adopted to smooth the trajectory, which will keep smoothing the trajectory until the difference between the current or the averaged coordinate and the next waypoint exceeds the given threshold.

### *Simulation to Real World Skill Transfer and Robot Motion Control*

With the demonstration data collected, stored, and analyzed in the previous steps, the next step involves determining how to use this data for robot motion control. One of the major obstacles is the coordinates misalignment between the DTEC environment and the robot control environment. The coordinates transformation matrix was calculated with the algorithm in Table 1. A total number of 728 readings from five locations (the four corners and centroid of the ceiling grid) are used. The average localization error is 0.0019 m, including 3DOF location and 4DOF quaternion orientation.

## Experimental Results and Discussion

The proposed cloud robotics framework and hierarchical robot learning model are evaluated for their improvement in skill transfer. The following evaluation metrics are adopted:

1) Learning model training time and error with different demonstration sources;
2) Robot task execution success rate with different demonstration sources;

**Table 1 DTEC-Gazebo Coordinates Transfer**

For each Rosbag file i=0,1,2,3,4:
    extract the Unity pose information for each stamp $p^{ut} = [x^{ut}, y^{ut}, z^{ut}, ox^{ut}, oy^{ut}, oz^{ut}, ow^{ut}]$
    remove outliers
    push back to a vector for batch processing
    For all extracted pose Unity $p^{ut}$:
        calculate real world pose: $p^{wt} = [x^{ut}, y^{ut}, z^{ut}, 0,0,0,1]$
        calculate transformation matrix $T^t$: $T^t = p^{wt-1} \cdot p^{ut}$
        push back $T^t$ to a vector and return its mean value of $T^i$
return $T^{cal} = mean(T^0 \sim T^4)$
For all $p^{ut}$:
    calculate $p^{wt\prime} = T^{cal} \cdot p^{ut}$
    calculate error: $e = p^{wt\prime} - p^{wt}$
return mean error

*Learning Model Performances*

Model training time and Mean Squared Error (MSE) loss are used as evaluation metrics as they are established practices from in federated learning research and robot HIL algorithms (Lin et al. 2022; Yang et al. 2020; Xie et al. 2020; Hayashi et al. 2022). Model training time shows how long it takes for the learning model to achieve convergence. In the context of this study, variations in training speed reflects the dataset quality since a consistent model architecture is employed (Yang et al. 2020). It also offers insights into the efficacy of introducing heterogeneous task demonstrations to enhance the quality of the cloud database. Second, MSE loss, are widely used for unsupervised learning models, wherein outputs comprise predictions or trajectories. It shows the average Euclidean distance between the demonstration and learning model generated trajectory (Xie et al., 2020; Lin et al. 2022). This computation results in the shown reduced learning errors and increased learning speed in Figure 9 and Table 2.

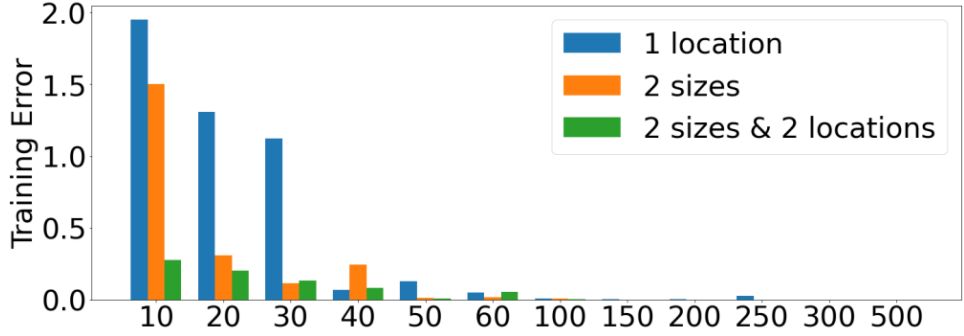
**Figure 9 Training Error Comparison for Three Datasets: 10-500 Epochs**

As can be seen in 9 and Table 2, for each epoch, the 1 location dataset (Dataset 1) often has the largest error. It shows the model learned from Dataset 1 generates a trajectory that is the most different from demonstrations, which cannot represent the human demonstrated skills. This is due to the small amount and homogeneity of data contained in this dataset (only 5 demonstrations: two by two ceiling tile installation in one location). Fsor the two other datasets, first, as they exist in the cloud database already, the workers do not need to make an extra effort to generate these demonstrations again. Second, with the increase of heterogeneous task demonstrations, the model learning error is largely reduced, especially with fewer training epochs.

Moreover, as the statistical features for different datasets in Table 2 show, Dataset 3 has the lowest mean value of training time and training error, illustrating the potent learning power provided by the heterogeneous data. In addition, the variance of the model performance from Dataset 3 is the lowest. As the variance shows how the number of training epochs affects the model performance, the small number suggests that using this dataset helps the model converge early (i.e. converge with fewer epochs) and be more robust towards noises.

**Table 2 Mean and Variance of Model Learning Speed and Errors**

|  | Training Time | | Training Error | |
|---|---|---|---|---|
|  | Mean | Variance | Mean | Variance |
| Dataset 1: 1 l size & 1 location | 19.356 | 405.142 | 0.389 | 0.454 |
| Dataset 2: 2 sizes & 1 locations | 19.264 | 399.528 | 0.184 | 0.183 |
| Dataset 3: 2 sizes & 2 locations | 19.115 | 389.370 | 0.0639 | 0.00885 |

*Robot Task Execution Performance*

The physical robot following the given motion primitives are shown in Figure 10.

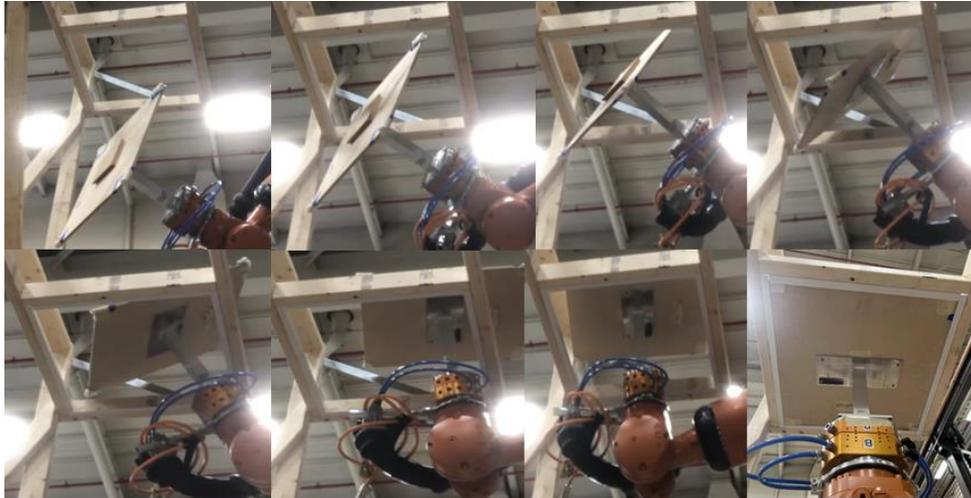

**Figure 10 A KUKA KR120 Robot Following Demonstrated Motion Primitives**

The evaluation framework aligns with the metrics used by Liang et al. (2020), wherein a 5 mm allowance between the tile's ultimate location and the grid is permissible. as shown in Figure 11, with successful installation instances (on the left) and cases that exceed the threshold (on the right).

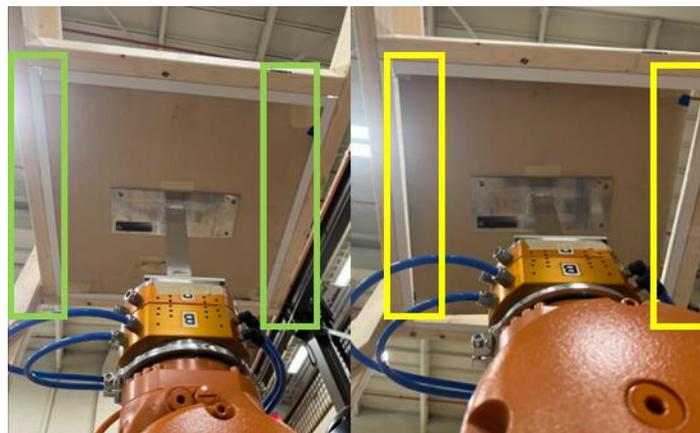

**Figure 11 Robot Task Execution Evaluation Criteria Example**

To avoid damaging the workpiece, the batch experiment with variables causing a lower success rate was repeated in Gazebo. The culmination of these evaluations is shown in Table 3, which enumerates the success rates across discrete experimental scenarios with five repetitions for each scenario. As can be seen in Table 3, when the robot only used the ad-hoc five demonstrations, it can successfully finish the given tasks 80% of the time. However, the robot cannot generalize this knowledge to the new task scenarios with a different target destination (will collide with the grid and not move to the designated target location) even with another round of training specially for the new locations. Yet, with the same training process (one for the original location and another for the new location), this problem is well addressed with five more demonstrations in new task scenarios in Dataset 3. The KUKA robot is capable of finishing installation tasks at both (2.95,1.5,2) and (3.5,1.5,2).

**Table 3 Simulated KUKA Robot Task Execution Performance Comparisons**

|  | Original location | New location | New location failed reasons |
|---|---|---|---|
| Dataset 1: 1 location | 80% | 0% | Robot collide with ceiling grid |
| Dataset 3: 2 locations | 100% | 80% | Unable to find valid motion plans |

*Limitations*

Admittedly, even accomplishing the research goals with experimental validations, the authors acknowledge that there are several limitations in this study:

First, when choosing the learning model input parameters, we present one set of working parameters composed of task parameters and object parameters. The reasons for such a choice are shown in the Research Methodology section. However, as this choice might not be the optimal one, future work can show how the alternative parameter settings improved the robot learning results and optimized the HIL model.

Second, the subtask proposed for the chosen case study is also one working way fitting into the proposed framework. There can be different ways to decompose one single task. For example, currently, the motion primitives are defined based on the pose, including location and orientation, of the task objects. Future work can explore the construction task decomposition variations to optimize the proposed hierarchical representation of construction installation tasks. one committed goal of this work is to reduce the workload of human workers.

Third, this current work only forms an open-loop control of construction robots. The data preprocessing, uploading to cloud, downloading to local, and edge learning are all in separate scripts and need to be connected by manual launch of several Linux scripts. Future work can explore more connected and advanced closed-loop construction robot learning and control schemes.

**Conclusions**

This paper targets several problems in construction robot application and craft skill programming. With previous research exploring the possibility of Imitation Learning to transfer the craft skills from human workers to robots, this work addressed the following problems:

1) The high physical workload in repetitive physical demonstrations: The high workload is first caused by the manipulation of heavy construction materials and second, by the need for repetitive demonstrations for new tasks. For the first reason, this study proposed to demonstrate in a high-fidelity VR and Digital Twin environment to avoid interacting with heavy materials. For the second reason, a cloud database is developed to connect the demonstration recording module to save past demonstrations and avoid demonstration repetitions.

2) The low generalizability in robot imitation learning: The majority robot IL models, including RL-based ones, are heavily dependent on environmental observations and task parameters. This study proposed a data-driven approach with a federated demonstration collection scheme and save crowdsourced data in the cloud database.

With more variation in the learning data, the learning model will naturally possess generalizable skills. The experiments with the simulated robot also validated this conclusion. In addition, by purposefully designing the demonstration to be manipulated workpiece trajectories, the learned trajectory can be generated to different robots, such as the Kuka KR120 prototyping robot.

3) The lack of learning model transparency and corresponding reduced trustworthiness: Traditional RL-based structure learns the craft skills in a black box. Human users, especially workers without programming expertise, will find it challenging to understand and trust robot learning models (Gunning et al. 2017). This study proposed the explainable hierarchical model which decodes craft skills as high-level sequential skills of subtasks, and low-level reactive skills that map the environmental observations to a certain choice of subtask. A generative model of VAE was used to model the reactive skills with very low error of averagely 0.000895 m with 500 epochs of training, while the GRU model determines sequence of actions with a mean error of 0. With both models achieving satisfactory performance, the craft skills are also decoded and recorded in interpretable robot learning models.

To summarize, this paper proposed a scalable scheme to program construction robots with craft skills. The proposed approach features cloud robotics and hierarchical learning to reduce the need for repetitive demonstrations and the workload for workers. Its enhanced explainability is also more user-friendly for workers without programming expertise.

**Acknowledgments**

The authors would like to acknowledge the financial support for this research received from the US National Science Foundation (NSF) FW-HTF 2025805 and FW-HTF 2128623. Any opinions and findings in this paper are those of the authors and do not necessarily represent those of the NSF. The authors would also like to acknowledge the contribution of Engineering Technician Justin Roelofs for demonstrating the ceiling installation process and building the wooden models used in the physical robot experiments.

Hsiao, K., & Lozano-Perez, T. (2006, October). Imitation learning of whole-body grasps. In *2006 IEEE/RSJ international conference on intelligent robots and systems* (pp. 5657-5662). IEEE.

Hua, J., Zeng, L., Li, G., & Ju, Z. (2021). Learning for a robot: Deep reinforcement learning, imitation learning, transfer learning. *Sensors*, *21*(4), 1278.

Hu, G., Tay, W. P., & Wen, Y. (2012). Cloud robotics: architecture, challenges and applications. *IEEE network*, *26*(3), 21-28.

Hussnain, A., Ferrer, B. R., & Lastra, J. L. M. (2018, May). Towards the deployment of cloud robotics at factory shop floors: A prototype for smart material handling. In *2018 IEEE Industrial Cyber-Physical Systems (ICPS)* (pp. 44-50). IEEE.

Jiao, Y., Wang, Y., Zhang, S., Li, Y., Yang, B., & Yuan, L. (2013). A cloud approach to unified lifecycle data management in architecture, engineering, construction and facilities management: Integrating BIMs and SNS. *Advanced Engineering Informatics*, *27*(2), 173-188.

Karp, P. D., Paley, S. M., & Greenberg, I. (1994, November). A storage system for scalable knowledge representation. In *Proceedings of the third international conference on Information and knowledge management* (pp. 97-104).

Kase, K., Paxton, C., Mazhar, H., Ogata, T., & Fox, D. (2020, May). Transferable task execution from pixels through deep planning domain learning. In *2020 IEEE International Conference on Robotics and Automation (ICRA)* (pp. 10459-10465). IEEE.

Kingma, Diederik P and Welling, Max. Auto-Encoding Variational Bayes. In The 2nd International Conference on Learning Representations (ICLR), 2013.

Kohler, M. D., Clayton, R. W., Bozorgnia, Y., Taciroglu, E., & Guy, R. (2022, December). The Community Seismic Network: Applications and Expansion to 1200 Stations. In *AGU Fall Meeting Abstracts* (Vol. 2022, pp. S42E-0195).

Kyrarini, M., Haseeb, M. A., Ristić-Durrant, D., & Gräser, A. (2019). Robot learning of industrial assembly task via human demonstrations. *Autonomous Robots*, *43*, 239-257.

Liu, C., S. Shirowzhan, S. M. E. Sepasgozar, and A. Kaboli. 2019. "Evaluation of classical operators and fuzzy logic algorithms for edge detection of panels at exterior cladding of buildings." *Buildings* 9 (2): 40. https://doi.org/10.3390/buildings9020040.

Liang, C. J., & Cheng, M. H. (2023). Trends in robotics research in occupational safety and health: a scientometric analysis and review. *International journal of environmental research and public health*, *20*(10), 5904.

Liang, C. J., Kamat, V. R., & Menassa, C. C. (2020). Teaching robots to perform quasi-repetitive construction tasks through human demonstration. *Automation in Construction*, *120*, 103370.

Liang, C. J., Wang, X., Kamat, V. R., & Menassa, C. C. (2021). Human–robot collaboration in construction: Classification and research trends. *Journal of Construction Engineering and Management*, *147*(10), 03121006.
32